\documentclass[sigconf]{acmart}

\AtBeginDocument{%
  }

\usepackage{graphicx}
\graphicspath{images/}
\usepackage{bm}
\usepackage{amsmath}
\usepackage{amsfonts}
\usepackage{multirow}
\usepackage{subfigure}

\usepackage{float}
\usepackage{color}

\begin{document}

\title{Orca: Ocean Significant Wave Height Estimation with Spatio-temporally Aware Large Language Models}

\settopmatter{authorsperrow=4}
\author{Zhe Li}
\email{zheli@stu.ecnu.edu.cn}
\affiliation{
  \institution{School of Data Science and Engineering, East China Normal University}
  \country{}
}

\author{Ronghui Xu}
\authornote{Corresponding author.}
\email{rhxu@stu.ecnu.edu.cn}
\affiliation{
  \institution{School of Data Science and Engineering, East China Normal University}
  \country{}
}

\author{Jilin Hu}
\email{hujilin@cs.aau.dk}
\affiliation{
    \institution{School of Data Science and Engineering, East China Normal University}
    \country{}
}
\author{Zhong Peng}
\email{zpeng@sklec.ecnu.edu.cn}
\affiliation{
  \institution{The State Key Laboratory of Estuarine and Coastal Research, ECNU}
  \country{}
}

\author{Xi Lu}
\email{xilu@stu.ecnu.edu.cn}
\affiliation{
  \institution{The State Key Laboratory of Estuarine and Coastal Research, ECNU}
  \country{}
}
\author{Chenjuan Guo}
\email{cjguo@dase.ecnu.edu.cn}
\affiliation{
  \institution{School of Data Science and Engineering, East China Normal University}
  \country{}
}
\author{Bin Yang}
\email{byang@dase.ecnu.edu.cn}
\affiliation{
  \institution{School of Data Science and Engineering, East China Normal University}
  \country{}
}

\renewcommand{\shortauthors}{Zhe Li et al.}

\begin{abstract}
Significant wave height (SWH) is a vital metric in marine science, and accurate SWH estimation is crucial for various applications, e.g., marine energy development, fishery, early warning systems for potential risks, etc. Traditional SWH estimation methods that are based on numerical models and physical theories are hindered by computational inefficiencies. Recently, machine learning has emerged as an appealing alternative to improve accuracy and reduce computational time. However, due to limited observational technology and high costs, the scarcity of real-world data restricts the potential of machine learning models. To overcome these limitations, we propose an ocean SWH estimation framework, namely Orca. 
Specifically, Orca enhances the limited spatio-temporal reasoning abilities of classic LLMs with a novel spatiotemporal aware encoding module. By segmenting the limited buoy observational data temporally, encoding the buoys' locations spatially, and designing prompt templates, Orca capitalizes on the robust generalization ability of LLMs to estimate significant wave height effectively with limited data. Experimental results on the Gulf of Mexico demonstrate that Orca achieves state-of-the-art performance in SWH estimation.
\end{abstract}

\keywords{Significant Wave Height, Large Language Model, Prompt Fine-tuning}

\maketitle

\section{Introduction}
Significant wave height (SWH) is an essential metric in marine science, reflecting the state of ocean activities.  
Since abnormal waves can cause extensive disruptions, leading to production losses, human casualties, and ecological damage, it is critical to accurately estimate the SWH and detect anomalies to ensure safety in a variety of applications, such as maritime navigation and marine energy development~\cite{Yang_Jin_Jia_Ye_2021, Woo_Park_2021, DBLP:journals/pvldb/PedersenYJ20,Bu_Park_Yu_Camps_2022,DBLP:conf/icde/KieuYGCZSJ22, DBLP:conf/icde/KieuYGJZHZ22,davidpvldb}. 
\begin{figure}[h]
    \centering
    \includegraphics[width=0.29\textwidth]{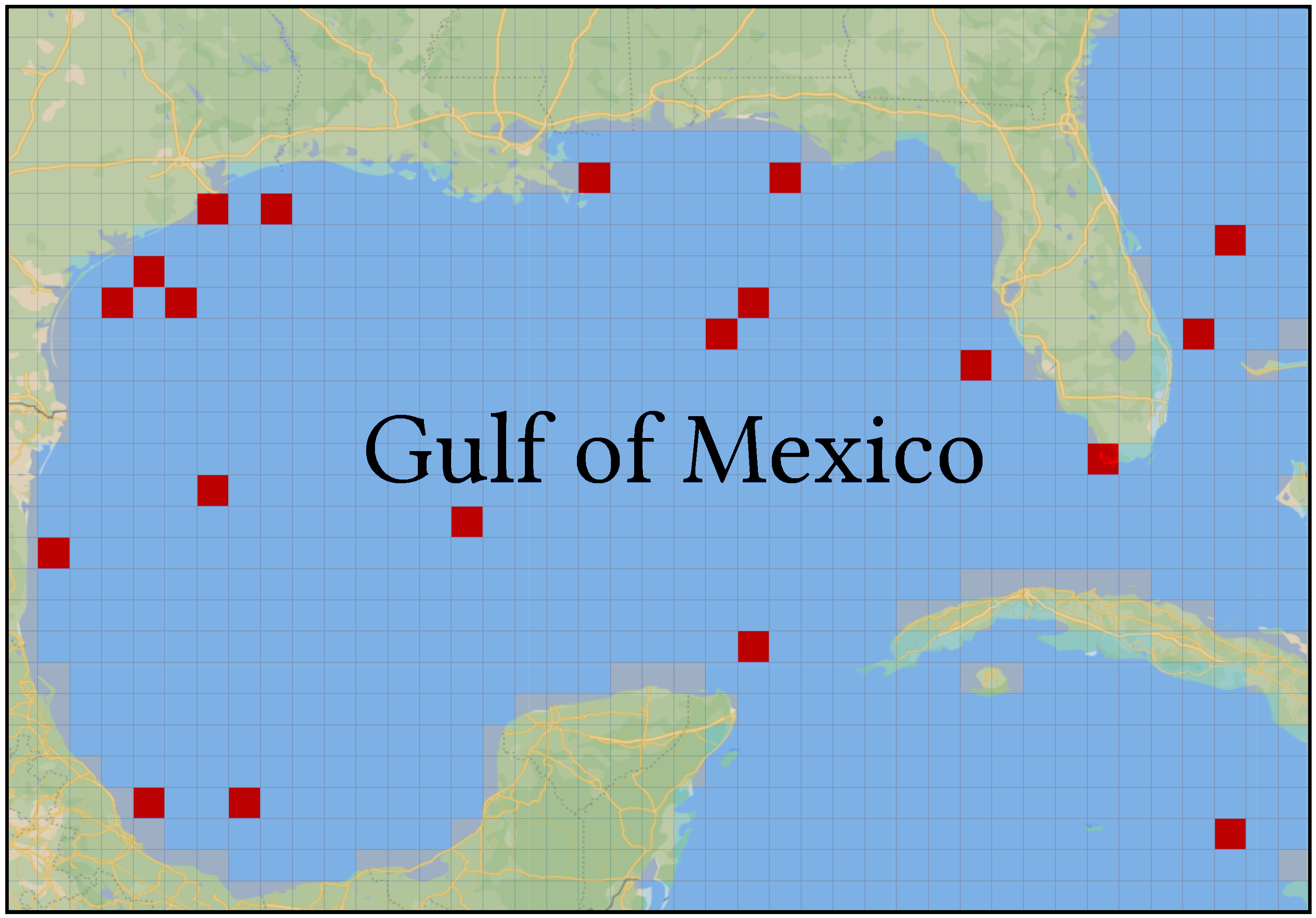}
    \vspace{-2ex}
    \caption{The buoy distribution in Gulf of Mexico.}
    \label{fig:map}
\end{figure}

Two categories of SWH estimation methods exist: traditional and machine learning-based methods. 
Specifically, traditional methods involve numerical models that simulate wave activities based on physical principles~\cite{Pierson_Neumann_James, ris1995spectral,hwang2021simulation}. 
Although these models have a robust theoretical foundation, they require extensive computational resources, which is too slow to estimate extremely high waves in time. Moreover, they are difficult to adapt to additional impact factors, which limits opportunities for performance improvement.  
Recently, machine learning-based time series methods~\cite{DBLP:journals/pvldb/QiuHZWDZGZJSY24,DBLP:journals/pacmmod/0002Z0KGJ23,DBLP:journals/pacmmod/Wu0ZG0J23,kaivldb24, yunyaovldb24,haoicde24,MileTS} have emerged as a promising alternative, offering improved accuracy and reduced computational time in SWH estimation~\cite{yang2024swhformer, Qi_Majda_2020, AnandJS23, wang2022significant}. 
However, they face two primary challenges. 
\textit{Firstly, the observed SWH data is too sparse.} Due to the limitations in observational technology~\cite{Qi_Majda_2020} and environmental factors, the deployed buoys in the ocean are very sparse. 
Figure~\ref{fig:map} shows the distribution of buoys in the Gulf of Mexico, where only the red grids are with buoys. 
Such scarcity hinders the development of machine learning-based methods.
\textit{Secondly, since a wave is a propagating dynamic disturbance of one or more quantities, the wave variations have strong spatio-temporal correlations. } Thus, it is crucial to capture such correlations in SWH estimation. However, current machine learning-based methods fail to capture such intricate relationships and discern the wave dynamic patterns implied in the data~\cite{Qi_Majda_2020, windspeed, QureshiKZU17}.

To address the abovementioned two challenges, we propose an ocean SWH estimation framework, namely \textit{Orca}. In recent years, Large Language Models~(LLMs) have shown remarkable performance in few-shot learning scenarios, matching the performance of task-specific models, e.g., arithmetic reasoning~\cite{Wei0SBIXCLZ22, KojimaGRMI22}, human mobility prediction~\cite{PromptCast}, and time series forecasting~\cite{pengiclr24, DBLP:journals/corr/abs-2402-02713}.
\textit{To tackle the issue of data sparsity in SWH estimation, we propose to use LLM as the backbone of estimation. }
Specifically, we invent a specific prompt templates and embedding module to leverage the pre-trained LLM for SWH estimation. With these designs, we aim to achieve accurate SWH estimations using limited observed data by leveraging the robust generalization capabilities of LLM. 
\textit{To enhance spatio-temporal reasoning, we segment buoy-based data into overlapping temporal patches, and propose a novel spatial encoding module. }In addition, we integrate additional information from the traditional numerical models by formulating a regularization term, such that we can anchor the model and establish scientific knowledge. Finally, our proposed model can not only overcome the limitations of traditional methods, but also improve computational efficiency and the precision of SWH predictions significantly. 

The study makes four main contributions. Firstly, we propose a framework for ocean SWH estimation based on LLMs with spatio-temporal awareness, Orca. Secondly, to tackle the data sparsity in SWH estimation, we invent a specific prompt templates and prompt embedding module to leverage LLM for estimation. Next, to enhance the spatio-temporal reasoning capabilities of LLMs, we propose a novel spatio-temporal aware encoding module to enable the detection of implicit wave dynamic patterns within the data. Finally, we conduct extensive experiments to demonstrate the computational efficiency and the accuracy of the proposed model.

\section{Preliminaries}
\noindent
\textbf{Significant Wave Height.}
Given a particular region and wave train, significant wave height~(SWH) is defined as the average height of the top one-third (1/3) waves. 

\noindent
\textbf{Buoy-based Data.}
Buoy-based data $\mathbf{X} \in \mathbb{R}^{F \times M \times T}$ encompasses the $F$ features collected by $M$ stationary oceanic buoys over $T$ continuous time intervals. 

\noindent
\textbf{Grid-based Significant Wave Height.}
Grid-based significant wave height (GSWH) $\mathbf{Y}\in \mathbb{R}^{K \times J \times T}$ records the average SWH within each grid area over $\mathrm{T}$ continuous time intervals, where $K$ and $J$ indicate the number of rows and columns, respectively.

\noindent
\textbf{Problem Definition.}
Given the buoy-based data $\mathbf{X}$ over $\mathrm{T}$ continuous time intervals, our goal is to estimate the GSWH values $\mathbf{Y}$ of $\mathrm{T}$ continuous time intervals.

\begin{figure*}[t]
    \centering
    \includegraphics[width=0.95\textwidth]{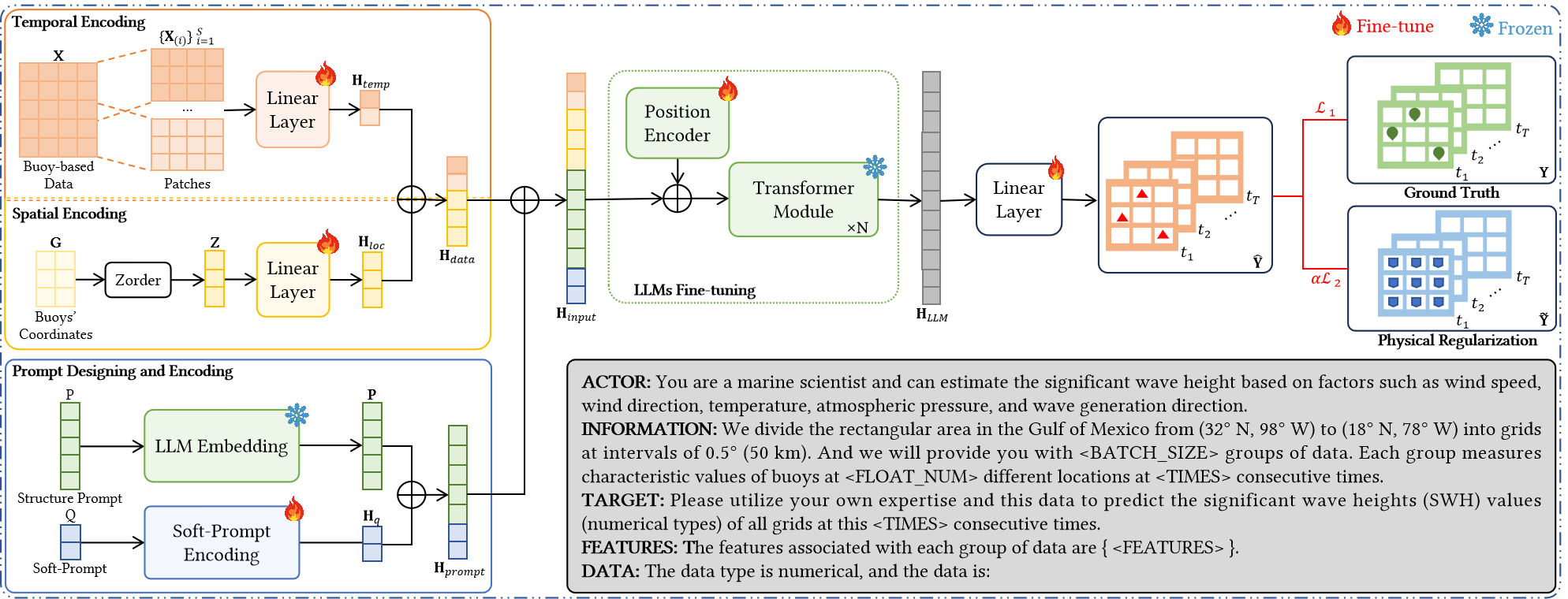}
    \vspace{-2ex}
    \caption{Overall framework of Orca.}
    \label{fig:model}
\end{figure*}

\section{Methodology}
\subsection{Prompt Designing and Encoding}
To guide LLMs towards an accurate comprehension of the data, we segment the prompt into five distinct components, as depicted in Figure ~\ref{fig:model}. ACTOR specifies the role of the LLMs; INFORMATION conveys the dimensions of the input data; TARGET clearly states the specific goals of the task; FEATURES specifies the semantics of the input data, such as wind direction (WDIR); and DATA declares the type of the input data, ensuring the LLMs accurately interprets numerical values rather than mistaking them for strings. 

The word embedding layer of LLMs takes input as the structured prompt $P=\{p_1, p_2, \ldots, p_E\}$, resulting in $\mathbf{P}=\{\mathbf{p}_1, \mathbf{p}_2, \ldots, \mathbf{p}_E\}$, where $E$ is the length of the prompt, $\mathbf{p}_i \in\mathbb{R}^D$ is the embedded vector of the corresponding token $p_i$, where $D$ is the dimension of the semantic space of LLMs.
 
Following P-Turning~\cite{liu2023gpt}, we randomly generate a fixed number of soft-prompt $Q=\{q_1, q_2, \ldots, q_R\}$, where $R$ is the length of soft-prompt. Then, ${Q}$ is fed into a pretrained embedding, obtaining $\mathbf{Q}=\{\mathbf{q}_1, \mathbf{q}_2, \ldots, \mathbf{q}_R\}$. We utilize learnable lightweight networks to model the dependency relationships between different embeddings:
\begin{equation}
    \mathbf{H}_q=\boldsymbol{\omega}_{1}(\mathrm{ReLU}(\boldsymbol{\omega}_{2}\mathrm{LSTM}(\mathbf{Q})+\boldsymbol{b}_{2}))+\boldsymbol{b}_{1},
\end{equation}
where $\boldsymbol{\omega}_{1}$, $\boldsymbol{b}_{1}$, $\boldsymbol{\omega}_{2}$, $\boldsymbol{b}_{2}$ are the trainable parameters of the prompt encoding module.

Finally, we concatenate $\mathbf{H}_q$ and $\mathbf{P}$ to improve the adaptability of the prompt to samples, obtaining the final prompt representations $\mathbf{H}_{prompt}=[\mathbf{H}_q; \mathbf{P}] \in\mathbb{R}^{(R+E)\times D}$.
 
\subsection{Spatio-temporal Encoding}
\textbf{Spatial Encoding.} As the phenomenon of wave variation at any specific maritime coordinate is inherently influenced by the wave movements in its vicinity, it is imperative to precise capture the spatial interconnections for both buoy and grid-based data. However, LLMs has limited spatial reasoning  capabilities. To overcome this challenge and effectively capture the spatial correlations essential, we introduce a spatial encoding module into our model.

Specifically,  we allocate buoys to their corresponding grid locations, and the locations of buoys are denoted as $\mathbf{G}=\{(\mathrm{u}_i, \mathrm{v}_i)\}_{i=1}^\mathrm{M}$, where $\mathrm{u}_i$ and $\mathrm{v}_i$ indicate the row and column grid indices for the $i$-th buoy. To effectively represent the spatial relationships within a multidimensional space, we employ the Z-order curve~\cite{DBLP:conf/vldb/LeeZLL07} to map the buoy coordinates to a one-dimensional binary embedding $\mathbf{Z}=\mathrm{Zorder}(\mathbf{G})$, $\mathbf{Z} \in \mathbb{R}^{M \times A}$. $\mathrm{Zorder}(\cdot)$ is the mapping function of the Z-order curve, and $A$ is defined by the number and size of grids. Subsequently, $\mathbf{Z}$ is input into a fully connected layer to generate the spatial embeddings $\mathbf{H}_{loc}=\mathrm{ReLU}(\boldsymbol{\omega}_{3}\mathbf{Z}+\boldsymbol{b}_{3})$, where $\boldsymbol{\omega}_{3}$, $\boldsymbol{b}_{3}$ are the trainable parameters of the spatial encoding module, and $\mathbf{H}_{loc}\in\mathbb{R}^\mathrm{M\times D}$ represents the spatial embeddings. This module ensures that our model accurately reflects the physical location of buoys in the estimations.

\textbf{Temporal Encoding.} To align the input buoy-based data with semantic space of LLMs and consider temporal information simultaneously, we redesign the input encoding layer, following a patch approach, which is used in time series analytics~\cite{nie2022time}. 

Specifically, the observed data $\mathbf{X}$ of buoys is segmented into overlapping patches $\mathbf{C}=\{\mathbf{X}_{(i)}\}_{i=1}^{S}$, where $\mathbf{X}_{(i)}\in\mathbb{R}^{F \times M \times L}$, ${S}=\lfloor\frac{T-L}{W}\rfloor+2$ is the number of patches, and $W$, $L$ are the stride and the patch length, respectively. We utilize a fully connected layer to align $\mathbf{C}$ with the semantic space of LLMs, formulated as:
\begin{equation}
    \mathbf{H}_{temp}=\mathrm{ReLU}(\boldsymbol{\omega}_4\mathbf{C}+\boldsymbol{b}_4),
\end{equation}
where $\mathbf{H}_{temp}\in\mathbb{R}^{S \times F \times M \times D}$ denotes the representation of the buoys' observed data, and $\boldsymbol{\omega}_4$, $\boldsymbol{b}_4$ are the trainable parameters.

We concatenate the representations aforementioned to derive the final input representation $\mathbf{H}_{input}=[\mathbf{H}_{prompt}; \mathbf{H}_{loc}; \mathbf{H}_{temp}] \in\mathbb{R}^\mathrm{ I \times F \times M \times D}$, and $I=R+E+1+S$.

\subsection{LLMs Fine-tuning}
To leverage the knowledge learned by LLMs during pre-training, we only fine-tune the positional encodings of the LLMs and freeze other parameters. We feed $\mathbf{H}_{input}$ into LLMs and acquire the output of the last latent layer:
\begin{equation}
\mathbf{H}_{LLM}=\mathrm{LLMs}(\mathbf{H}_{input}).
\end{equation}
Here, $\mathbf{H}_{LLM} \in  \mathbb{R}^{I \times F \times M \times D}$ is the output of LLMs. We perform average pooling on $\mathbf{H}_{LLM}$ to aggregate the feature dimension:
\begin{equation}
\mathbf{H}_{pool}=\mathrm{AvgPool}(\mathbf{H}_{LLM}).
\end{equation}
$\mathbf{H}_{pool}$ is then flattened to a ($I \times M \times D$)-dimensional embedding. We utilize a fully connected layer and a reshape function to obtain the estimated GSWH values:

\begin{equation}\mathbf{\widehat{Y}}=\mathrm{Reshape}(\boldsymbol{\omega}_{5}\mathbf{H}_{pool}+\boldsymbol{b}_{5}),
\end{equation}
where $\mathrm{Reshape}(\cdot)$ is a function that reshapes a ($K \times J \times T$)-dimensional embedding to estimated GSWH values $\mathbf{\widehat{Y}} \in \mathbb{R}^{K \times J \times T}$. $\boldsymbol{\omega}_{5}$, $\boldsymbol{b}_{5}$ are the weight and the bias of the layer, respectively.

\subsection{Optimizing with Physical Regularization}

We minimize the difference between the observed SWH values from buoys and the estimated GSWH values. The loss function is:
\begin{equation}
\mathcal{L}_1=\frac{1}{M}\sum_{i=1}^{M}(\mathrm{Y}_{u_i,v_i}-\hat{\mathrm{Y}}_{u_i,v_i})^2,
\end{equation}
where $M$ is the number of buoys, $(u_i,v_i)$ is are the $2D$ coordinates of the $i$-th buoy, $\mathrm{Y}_{u_i,v_i}$ denotes the observed SWH values of the $i$-th buoy, and $\hat{\mathrm{Y}}_{u_i,v_i}$ indicates the estimated SWH values of our proposed model.

To guide the model to follow physcial principles, we utilize data from the numerical model to formulate a regularization term:
\begin{equation}
\mathcal{L}_2=\frac{1}{K \times J}\sum_{i=1}^{K}\sum_{j=1}^{J}(\tilde{\mathrm{Y}}_{i,j}-\hat{\mathrm{Y}}_{i,j})^2,
\end{equation} 
where $\tilde{\mathrm{Y}}_{i,j}$ represents the GSWH value,  generated by the numerical model, for the grid at the $i$-th row and $j$-th column, $\hat{\mathrm{Y}}_{i,j}$ is the corresponding estimated value produced by our model.

Overall, the training objective can be formalized as:
\begin{equation}
\mathcal{L}=\mathcal{L}_1+\alpha\mathcal{L}_2,
\end{equation}
where hyperparameter $\alpha$ is the weight of $\mathcal{L}_2$. 

\makeatletter
\newcommand\figcaption{\def\@captype{figure}\caption}
\newcommand\tabcaption{\def\@captype{table}\caption}
\makeatother
\begin{figure*}
    \centering
    \begin{minipage}{0.33\textwidth}
        \centering
        \vspace{-4ex}
        \tabcaption{Accuracy (best values in bold).}
        \vspace{1ex}
        \resizebox{0.8\textwidth}{!}{
        \renewcommand\arraystretch{1.7}
        \begin{tabular}{c|c|c|c}
            \hline
            {\color[HTML]{000000} \textbf{Model}} & {\color[HTML]{000000} \textbf{MAE}} & {\color[HTML]{000000} \textbf{MSE}} & \textbf{RMSE}   \\ \hline
            \textbf{GWD~\cite{center2014user}}         & 0.3949          & 0.2000          & 0.4472   \\ \hline
            \textbf{PatchTST~\cite{nie2022time}}       & 0.6652          & 0.9375          & 0.9682   \\ \hline
            \textbf{GPT-2~\cite{radford2019language}}  & 0.3962          & 0.2063          & 0.4542   \\ \hline
            \textbf{GPT4TS~\cite{zhou2023one}}   & 0.3692          & 0.1796          & 0.4238   \\ \hline
            \textbf{Orca}                              & \textbf{0.2372} & \textbf{0.0838} & \textbf{0.2895} \\
            \hline
        \end{tabular}
        }
        \label{table:result}
    \end{minipage}
    \begin{minipage}{0.33\textwidth}
        \centering
        \includegraphics[width=0.95\linewidth]{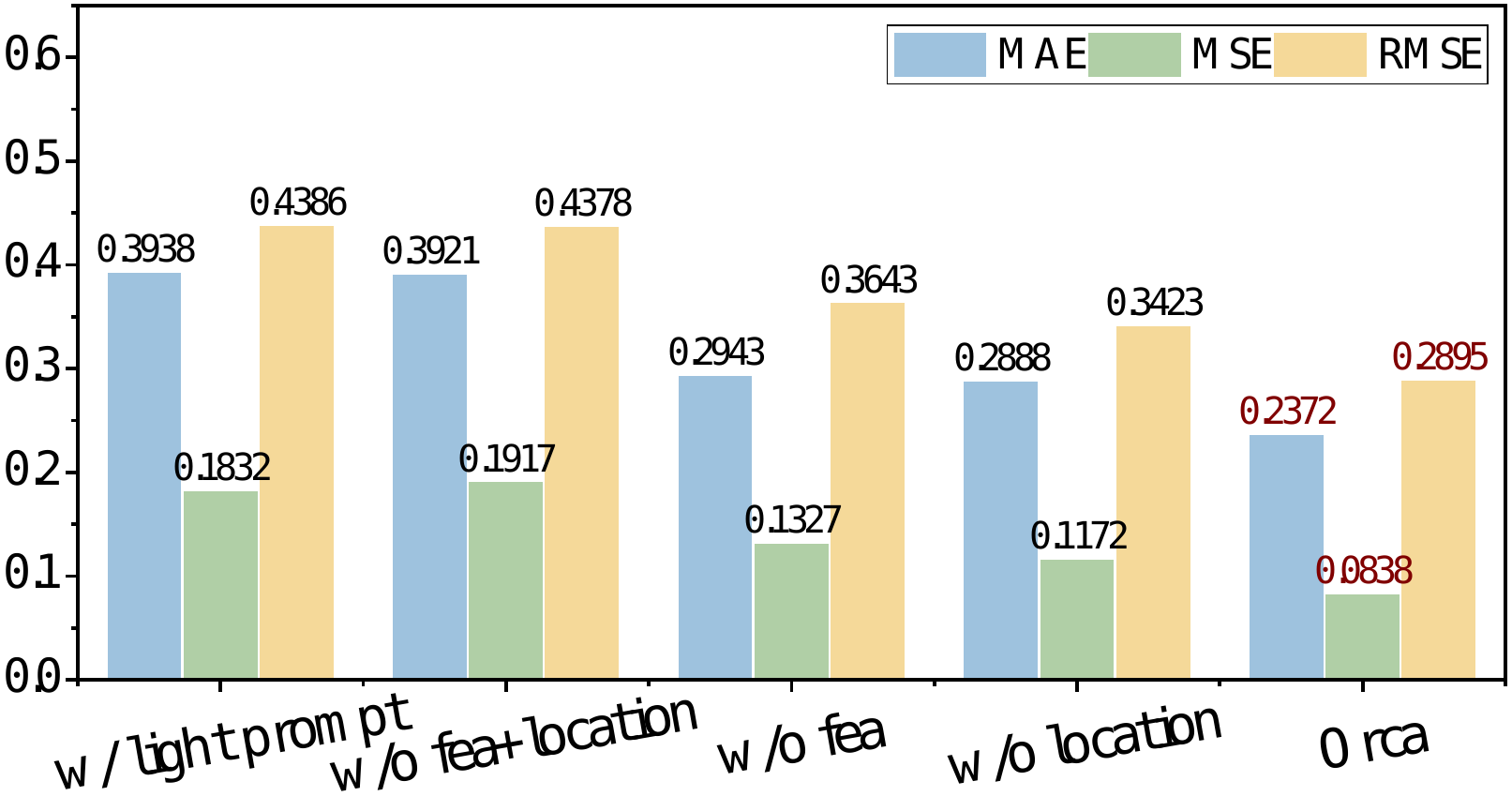}
        \vspace{-2ex}
        \figcaption{Ablation Studies.}
        \label{fig:ablation}
    \end{minipage}
    \begin{minipage}{0.33\textwidth}
        \centering
        \includegraphics[width=0.9\textwidth]{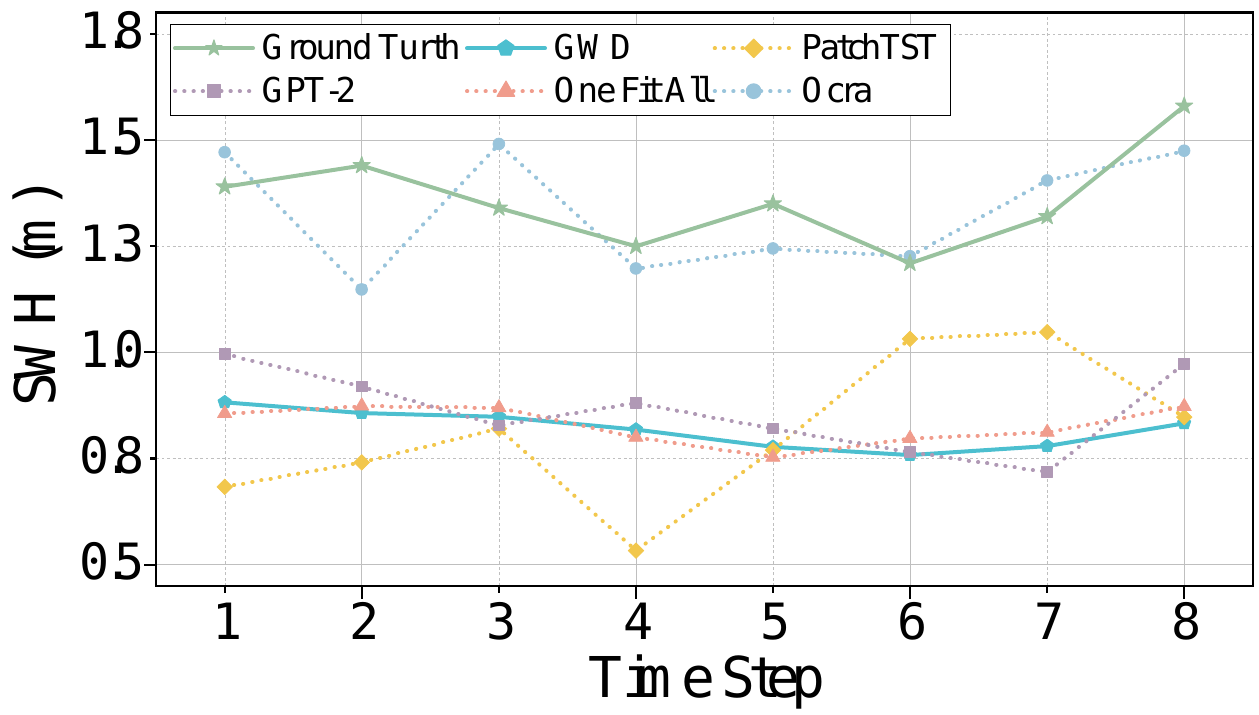}
        \vspace{-2ex}
        \figcaption{Estimated SWH at buoy 42040.}
        \label{fig:42040}
    \end{minipage}
\end{figure*}

\section{Experiments}
\textbf{Datasets and Baselines.}
In this paper, we focus on a defined rectangular region within the Gulf of Mexico, spanning from 32°N to 18°N latitude and 98°W to 78°W longitude. This area is uniformly segmented into $29\times41$ grids, with each grid cell representing a 0.5° (approximately 50 km) square. Then, the Buoy-based data is from the National Data Buoy Center dataset~(NDBC)\footnote{https://www.ndbc.noaa.gov/}, which can provide key measurements such as surface wind speed, sea surface temperature, and significant wave height. The corresponding numerical model data is from the Global Wave Database (GWD), provided by the Estuary and Coastal Laboratory\footnote{http://www.sklec.ecnu.edu.cn/}, collects wave simulation data generated by version 4.18 of the WaveWatch III~\cite{center2014user}. 

We compare Orca with various competitive baseline models including a traditional method using physical principles GWD~\cite{center2014user}, a SOTA time series forecasting method PatchTST~\cite{nie2022time}, a large language model without spatio-temporal aware encoding GPT-2~\cite{radford2019language}, and a LLM based time series forecasting method GPT4TS~\cite{zhou2023one}.

\textbf{Experiment Settings.}
All the experiments are conducted on a high-performance NVIDIA RTX3090 24GB GPU. 
We utilize  GPT-2~\cite{radford2019language} as the backbone network for our model.
The training process is executed over a maximum of 50 epochs. 
The learning rate and $\alpha$ are set at 0.001 and 0.3, respectively. We split NDBC dataset into non-overlapping  train, validation, and test sets with a ratio of 8:1:1. The GWD dataset is only used for training. The time interval is set to 3-hour for both datasets. We conduct the evaluation using buoy-based data from the test set as the ground truth, and employ MSE, RMSE, and MAE as metrics.

\textbf{Performance comparison.}
Table~\ref{table:result} shows the comparative results of various models on SWH estimation. Time series models, specifically PatchTST, exhibit limitations in capturing spatial relationships and dynamic wave patterns, particularly in data-scarce environments, leading to poor performance. Conversely, applying GPT-2 directly or fine-tuning it with GPT4TS approach allows LLMs to achieve results comparable to the numerical model GWD. This validates the efficacy of LLMs in few-shot learning scenarios. Notably, Orca achieves the best performance, affirming the superior impact of spatio-temporal encoding module and designed prompts in refining LLMs for SWH estimation.

\textbf{Ablation Studies.}
To verify the impact of Orca’s key components, we conduct ablation studies, exploring the following variants:
(1) w/ light prompt: Orca utilizing a light prompt: ``ACTOR: You are a marine scientist. TARGET: I will provide you with several sets of data describing the marine environment. Please utilize your own expertise and this data to predict the significant wave heights.''
(2) w/o fea+location: It excludes both the FEATURES from the prompt and the spatial encoding module within Orca.
(3) w/o fea: It removes FEATURES in prompt.
(4) w/o location: It removes the spatial encoding module.

Figure~\ref{fig:ablation} shows that different variants yield distinct performance. The performance of w/ light prompt, while worse than Ocra, outperforms the numerical model GWD, indicating that prompt fine-tuning can significantly enhance the generalization capabilities of LLMs for this task. Moreover, the removal of FEATURES or the spatial encoding module from Orca (i.e., w/o fea+location, w/o fea, and w/o location) results in a noticeable decrease in performance. This decline confirms the vital role of both the designed prompts and the spatial encoding module in achieving accurate SWH estimations, underscoring their indispensability for the model’s success.

\textbf{Time Efficiency Analysis.}
The time efficiency of the traditional model for SWH estimation is notably low, taking about 4 minutes to estimate one day’s SWH on a high-performance 64-core computer. In contrast, Orca showcases remarkable computational efficiency, completing short-term (1 day) estimation in just 0.0095 seconds. Even for medium-term (8 days) and long-term (28 days) estimations, Orca remains this efficiency, taking only 5.5527 seconds and 27.3071 seconds, respectively. Thus, Orca is significantly more computationally efficient than traditional models in SWH estimation.

\textbf{Visualization.}
We evaluate the estimated SWH from various models over 8 consecutive time steps at buoy 42040, comparing them with the ground truth (see Figure~\ref{fig:42040}). The estimations of Orca closely match the ground truth, effectively capturing notable fluctuations. In contrast, since the modeling of GWD is  limited to fixed factors, its trend is overly smooth, failing to capture significant fluctuations.
Similarly, other baseline models tend to underestimate SWH values, and fail to  accurately capture wave fluctuations.

\begin{figure}[t]
    \centering
    \subfigure[GWD]{
    \includegraphics[width=0.225\textwidth]{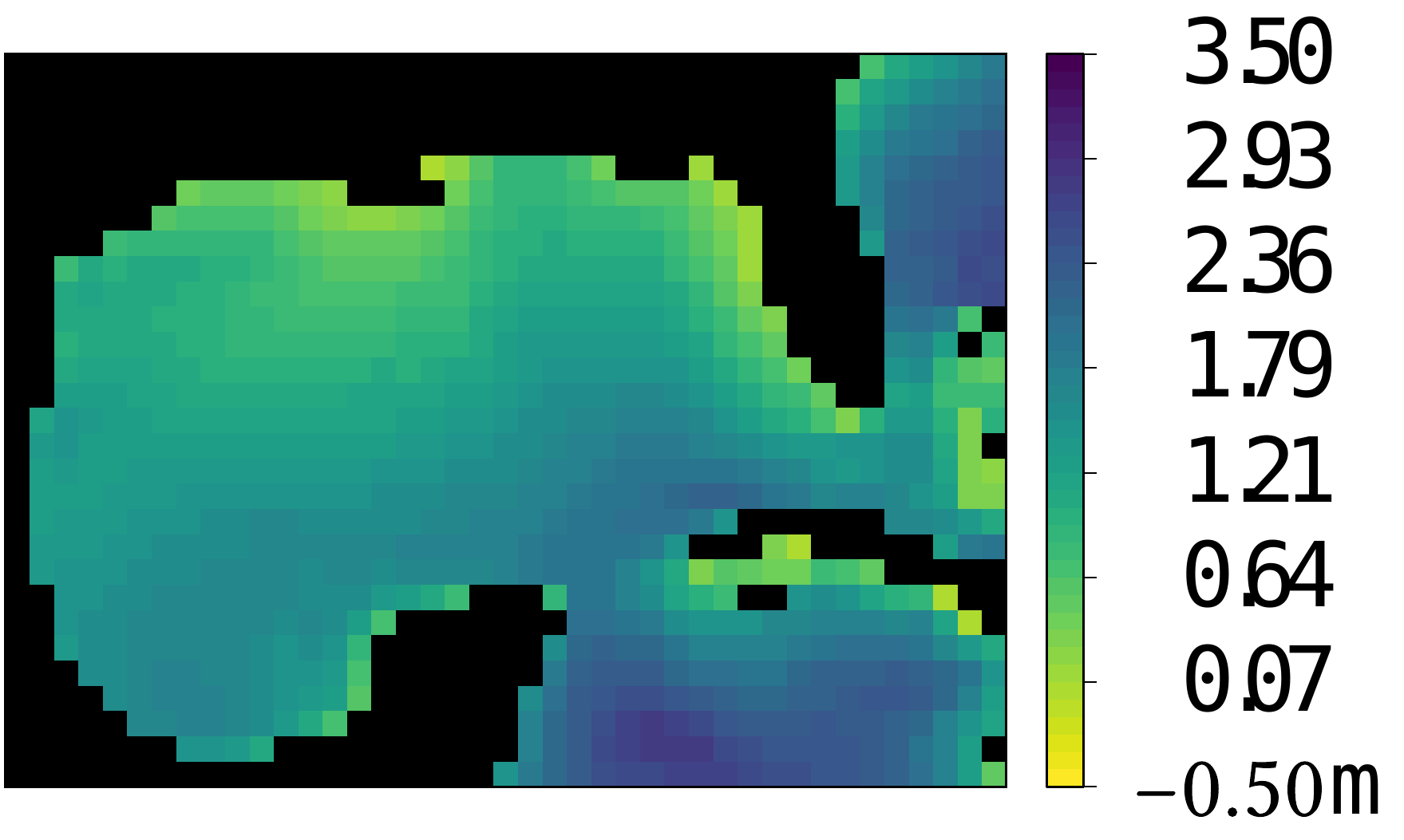}
        \label{grid.GWD}
    }
    \subfigure[Orca]{
        \includegraphics[width=0.225\textwidth]{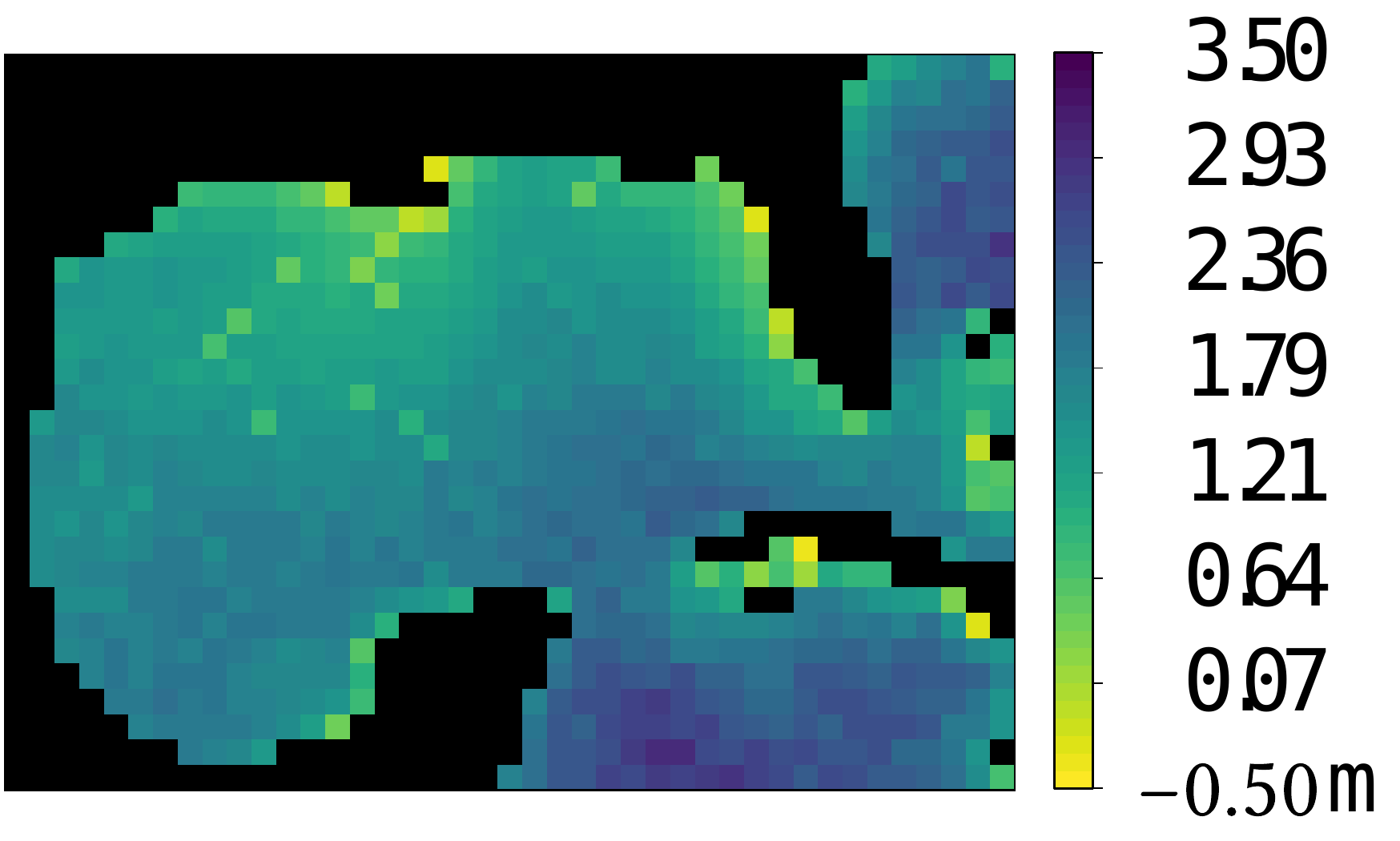}
        \label{grid.Orca}
    }
    \vspace{-1.5em}
    \caption{The heat map of GSWH at a certain moment.}
    \label{fig:grid}
    \vspace{-2em}
\end{figure}

We use heat maps to show GSWH distribution in the target sea area at a certain moment (see Figure~\ref{fig:grid}). Each grid cell presents a GSWH value. The data reveals that wave activity intensifies with increasing distance from the coast. Orca is not only able to capture similar results as the numerical model GWD, but also more pronounced fluctuations, further confirming the differences between the two models in Figure \ref{fig:42040}. 

\section{Conclusions}
This paper presents a solution to the significant wave height (SWH) estimation problem in scenarios of limited data. Firstly, we introduce LLMs as the backbone to leverage their few-shot learning ability. We design a specific prompt templates and embedding module to adapt LLMs to the SWH estimation task. Secondly, we segment buoy-based data into overlapping temporal patches, and employ a novel spatial encoding module to improve spatio-temporal reasoning. Extensive experiments confirm the effectiveness of our proposed model, Orca. In the future, we aim to boost Orca’s efficiency, facilitating both one-step and multi-step SWH predictions.

\begin{acks}
 This work was partially supported by National Natural Science Foundation of China (62372179).
\end{acks}

\bibliographystyle{unsrt}
\balance
\bibliography{sample-base}

\begin{thebibliography}{10}

\bibitem{Yang_Jin_Jia_Ye_2021}
Shuai Yang, Shuanggen Jin, Yan Jia, et~al.
\newblock Significant wave height estimation from joint {CYGNSS} {DDMA} and {LES} observations.
\newblock {\em Sensors}, 21(18):6123, 2021.

\bibitem{Woo_Park_2021}
Hye~Jin Woo and Kyung~Ae Park.
\newblock Estimation of extreme significant wave height in the northwest pacific using satellite altimeter data focused on typhoons.
\newblock {\em Remote Sensing}, 13(6):1063, 2021.

\bibitem{DBLP:journals/pvldb/PedersenYJ20}
Simon~Aagaard Pedersen, Bin Yang, and Christian~S. Jensen.
\newblock Anytime stochastic routing with hybrid learning.
\newblock {\em Proc. {VLDB} Endow.}, 13(9):1555--1567, 2020.

\bibitem{Bu_Park_Yu_Camps_2022}
Jinwei Bu, Hyuk Park, Kegen Yu, et~al.
\newblock Estimation of significant wave height using the features of cygnss delay doppler map.
\newblock In {\em Proceedings of the International Geoscience and Remote Sensing Symposium}, 2022.

\bibitem{DBLP:conf/icde/KieuYGCZSJ22}
Tung Kieu, Bin Yang, Chenjuan Guo, Razvan{-}Gabriel Cirstea, Yan Zhao, Yale Song, and Christian~S. Jensen.
\newblock Anomaly detection in time series with robust variational quasi-recurrent autoencoders.
\newblock In {\em {ICDE}}, pages 1342--1354, 2022.

\bibitem{DBLP:conf/icde/KieuYGJZHZ22}
Tung Kieu, Bin Yang, Chenjuan Guo, Christian~S. Jensen, Yan Zhao, Feiteng Huang, and Kai Zheng.
\newblock Robust and explainable autoencoders for unsupervised time series outlier detection.
\newblock In {\em {ICDE}}, pages 3038--3050, 2022.

\bibitem{davidpvldb}
David Campos, Tung Kieu, Chenjuan Guo, Feiteng Huang, Kai Zheng, Bin Yang, and Christian~S. Jensen.
\newblock Unsupervised time series outlier detection with diversity-driven convolutional ensembles.
\newblock {\em Proc. {VLDB} Endow.}, 15(3):611--623, 2022.

\bibitem{Pierson_Neumann_James}
Willard~J. Pierson, Gerhard Neuman, and Richard~W. James.
\newblock Practical methods for observing and forecasting ocean waves by means of wave spectra and statistics.
\newblock 1955.

\bibitem{ris1995spectral}
R.~C. Ris, L.~H. Holthuijsen, and N.~Booij.
\newblock {A} spectral model for waves in the near shore zone.
\newblock {\em Coastal Engineering}, pages 68--78, 1995.

\bibitem{hwang2021simulation}
Paul~A. Hwang, Jeffrey~D. Ouellette, Jakov~V. Toporkov, et~al.
\newblock {A} simulation study of significant wave height retrieval from bistatic scattering of signals of ppportunity.
\newblock {\em Transactions on Geoscience and Remote Sensing}, 19:1--5, 2021.

\bibitem{DBLP:journals/pvldb/QiuHZWDZGZJSY24}
Xiangfei Qiu, Jilin Hu, Lekui Zhou, Xingjian Wu, Junyang Du, Buang Zhang, Chenjuan Guo, Aoying Zhou, Christian~S. Jensen, Zhenli Sheng, and Bin Yang.
\newblock {TFB:} towards comprehensive and fair benchmarking of time series forecasting methods.
\newblock {\em Proc. {VLDB} Endow.}, 17(9):2363--2377, 2024.

\bibitem{DBLP:journals/pacmmod/0002Z0KGJ23}
David Campos, Miao Zhang, Bin Yang, Tung Kieu, Chenjuan Guo, and Christian~S. Jensen.
\newblock Light{TS}: Lightweight time series classification with adaptive ensemble distillation.
\newblock {\em Proc. {ACM} Manag. Data}, 1(2):171:1--171:27, 2023.

\bibitem{DBLP:journals/pacmmod/Wu0ZG0J23}
Xinle Wu, Dalin Zhang, Miao Zhang, Chenjuan Guo, Bin Yang, and Christian~S. Jensen.
\newblock Auto{CTS}+: Joint neural architecture and hyperparameter search for correlated time series forecasting.
\newblock {\em Proc. {ACM} Manag. Data}, 1(1):97:1--97:26, 2023.

\bibitem{kaivldb24}
Kai Zhao, Chenjuan Guo, Peng Han, Miao Zhang, Yunyao Cheng, and Bin Yang.
\newblock Multiple time series forecasting with dynamic graph modeling.
\newblock {\em Proc. {VLDB} Endow.}, 2024.

\bibitem{yunyaovldb24}
Yunyao Cheng, Peng Chen, Chenjuan Guo, Kai Zhao, Qingsong Wen, Bin Yang, and Christian~S. Jensen.
\newblock Weakly guided adaptation for robust time series forecasting.
\newblock {\em Proc. {VLDB} Endow.}, 2024.

\bibitem{haoicde24}
Hao Miao, Yan Zhao, Chenjuan Guo, Bin Yang, Zheng Kai, Feiteng Huang, Jiandong Xie, and Christian~S. Jensen.
\newblock A unified replay-based continuous learning framework for spatio-temporal prediction on streaming data.
\newblock {\em {ICDE}}, 2024.

\bibitem{MileTS}
Razvan-Gabriel Cirstea, Bin Yang, and Chenjuan Guo.
\newblock Graph attention recurrent neural networks for correlated time series forecasting.
\newblock In {\em {MileTS19@KDD}}, 2019.

\bibitem{yang2024swhformer}
Zhiding Yang and Weimin Huang.
\newblock Swhformer: {A} vision transformer for significant wave height estimation from nautical radar images.
\newblock {\em Transactions on Geoscience and Remote Sensing}, 62:1--13, 2024.

\bibitem{Qi_Majda_2020}
Di~Qi and Andrew~J. Majda.
\newblock Using machine learning to predict extreme events in complex systems.
\newblock {\em Proceedings of the National Academy of Sciences}, page 52–59, 2020.

\bibitem{AnandJS23}
Pritam Anand, Shantanu Jain, and Harsh Savaliya.
\newblock New improved wave hybrid models for short-term significant wave height forecasting.
\newblock {\em {IEEE} Access}, 11:109841--109855, 2023.

\bibitem{wang2022significant}
Changyang Wang, Kegen Yu, Kefei Zhang, et~al.
\newblock Significant wave height retrieval based on multivariable regression models developed with {CYGNSS} data.
\newblock {\em Transactions on Geoscience and Remote Sensing}, 61:1--15, 2022.

\bibitem{windspeed}
T.G. Barbounis and J.B. Theocharis.
\newblock Locally recurrent neural networks optimal filtering algorithms: application to wind speed prediction using spatial correlation.
\newblock In {\em Proceedings of the International Joint Conference on Neural Networks}, volume~5, pages 2711--2716, 2005.

\bibitem{QureshiKZU17}
Aqsa~Saeed Qureshi, Asifullah Khan, Aneela Zameer, et~al.
\newblock Wind power prediction using deep neural network based meta regression and transfer learning.
\newblock {\em Applied Soft Computing}, 58:742--755, 2017.

\bibitem{Wei0SBIXCLZ22}
Jason Wei, Xuezhi Wang, Dale Schuurmans, et~al.
\newblock Chain-of-thought prompting elicits reasoning in large language models.
\newblock In {\em Proceesings of the Advances in Neural Information Processing Systems}, 2022.

\bibitem{KojimaGRMI22}
Takeshi Kojima, Shixiang~Shane Gu, Machel Reid, et~al.
\newblock Large language models are zero-shot reasoners.
\newblock In {\em Proceedings of the Advances in Neural Information Processing Systems}, 2022.

\bibitem{PromptCast}
Hao Xue and Flora~D. Salim.
\newblock Promptcast: A new prompt-based learning paradigm for time series forecasting.
\newblock {\em {IEEE} Transactions on Knowledge and Data Engineering}, pages 1--14, 2023.

\bibitem{pengiclr24}
Peng Chen, Yingying Zhang, Yunyao Cheng, Yang Shu, Yihang Wang, Qingsong Wen, Bin Yang, and Chenjuan Guo.
\newblock Pathformer: Multi-scale transformers with adaptive pathways for time series forecasting.
\newblock {\em {ICLR}}, 2024.

\bibitem{DBLP:journals/corr/abs-2402-02713}
Ming Jin, Yifan Zhang, Wei Chen, Kexin Zhang, Yuxuan Liang, Bin Yang, Jindong Wang, Shirui Pan, and Qingsong Wen.
\newblock Position paper: What can large language models tell us about time series analysis.
\newblock {\em CoRR}, abs/2402.02713, 2024.

\bibitem{liu2023gpt}
Xiao Liu, Yanan Zheng, Du~Zhengxiao, et~al.
\newblock {GPT} understands, too.
\newblock {\em AI Open}, 2023.

\bibitem{DBLP:conf/vldb/LeeZLL07}
Ken C.~K. Lee, Baihua Zheng, Huajing Li, et~al.
\newblock Approaching the skyline in {Z} order.
\newblock In {\em Proceedings of the International Conference on Very Large Data Bases}, pages 279--290, 2007.

\bibitem{nie2022time}
Yuqi Nie, Nam~H. Nguyen, Phanwadee Sinthong, et~al.
\newblock A time series is worth 64 words: Long-term forecasting with transformers.
\newblock In {\em Proceedings of the International Conference on Learning Representations}, 2023.

\bibitem{center2014user}
Environmental~Modeling Center.
\newblock User manual and system documentation of {WAVEWATCH} {III} {R} version 4.18.
\newblock 2014.

\bibitem{radford2019language}
Alec Radford, Jeffrey Wu, Rewon Child, et~al.
\newblock Language models are unsupervised multitask learners.
\newblock {\em OpenAI Blog}, 1(8):9, 2019.

\bibitem{zhou2023one}
Tian Zhou, Peisong Niu, Xue Wang, Liang Sun, et~al.
\newblock One fits all: Power general time series analysis by pretrained {LM}.
\newblock In {\em Proceesings of the Advances in Neural Information Processing Systems}, 2023.

\end{thebibliography}

\end{document}